\documentclass{ifacconf}

\usepackage{graphicx}      
\usepackage{natbib}        

\usepackage{amsmath}
\usepackage{amsfonts}
\usepackage{graphicx}
\usepackage{mathrsfs}
\usepackage{bbm}
\usepackage{amssymb}
\usepackage{rotating}
\usepackage{comment}
\usepackage{xcolor}
\usepackage{soul}
\usepackage{subcaption}
\usepackage{amssymb} 

\newtheorem{definition}{Definition}[section]

\usepackage{xcolor}
\usepackage{soul}

\makeatletter

\renewcommand\hl{%
  \let\bgroup{\tikz[baseline,anchor=base]%
    \bgroup\aftergroup\endtikzpicture}%
  \egroup
}
\makeatother

\definecolor{lightblue}{rgb}{0.85, 0.92, 1.0}
\definecolor{lightgray}{rgb}{0.93, 0.93, 0.93}
\definecolor{darkgreen}{rgb}{0.44, 0.62, 0.35}

\begin{document}
\begin{frontmatter}

\title{Multi-layer Abstraction for Nested Generation of Options (MANGO) in Hierarchical Reinforcement Learning} 


\author[hit]{Alessio Arcudi$^\#$} 
\author[dei]{Davide Sartor$^\#$} 
\author[hit]{Alberto Sinigaglia}
\author[ams]{Vincent Fran\c{c}ois-Lavet}
\author[hit,dei]{Gian Antonio Susto}

\address[hit]{Human Inspired Technology Research Center, Università di Padova, Padova, PD 35121 IT.}
\address[dei]{Dep. of Information Engineering, Università di Padova, Padova, PD 35121 IT.}
\address[ams]{Vrije Universiteit Amsterdam, Amsterdam, Netherlands}

\begin{abstract}
This paper introduces MANGO (Multilayer Abstraction for Nested Generation of Options), a novel hierarchical reinforcement learning framework designed to address the challenges of long-term sparse reward environments. 
MANGO decomposes complex tasks into multiple layers of abstraction, where each layer defines an abstract state space and employs options to modularize trajectories into macro-actions. These options are nested across layers, allowing for efficient reuse of learned movements and improved sample efficiency. 
The framework introduces intra-layer policies that guide the agent's transitions within the abstract state space, and task actions that integrate task-specific components such as reward functions. Experiments conducted in procedurally-generated grid environments demonstrate substantial improvements in both sample efficiency and generalization capabilities compared to standard RL methods. 
MANGO also enhances interpretability by making the agent's decision-making process transparent across layers, which is particularly valuable in safety-critical and industrial applications. 
Future work will explore automated discovery of abstractions and abstract actions, adaptation to continuous or fuzzy environments, and more robust multi-layer training strategies.
\end{abstract}


\end{frontmatter}

\def\thefootnote{\#}\footnotetext{Equal contribution. Corresponding author: Davide Sartor (email: davide.sartor.4@phd.unipd.it). Authors' email: {alberto.sinigaglia, alessio.arcudi}@phd.unipd.it, vincent.francoislavet@vu.nl, gianantonio.susto@unipd.it.}

\renewcommand{\thefootnote}{\arabic{footnote}}


\section{Introduction}

Reinforcement Learning (RL) has proved immensely successful in tackling a variety of decision-making tasks, yet its efficacy can diminish in environments characterized by large state spaces and sparse rewards. As the number of possible states expands and the temporal horizon grows, standard RL methods often struggle to identify effective strategies. 
This challenge has spurred the development of Hierarchical RL (HRL) approaches, which divide complex tasks into smaller, more manageable subtasks \citep{Dietterich1999HierarchicalRL,sutton1999between,hengst2002discovering}.

Hierarchy in artificial intelligence has been studied for decades \citep{Sacerdoti1974PlanningIA}. 
A central concept in HRL involves decomposing an overarching Markov Decision Process (MDP) into multiple, smaller MDPs \citep{sutton1999between,Silver2012CompositionalPU,Konidaris2009EfficientSL}. 
These sub-MDPs capture subtasks defined by core states or transitions of interest. \cite{Ho2019TheVO,Ho2021PeopleCS} further clarify hierarchy in RL by distinguishing between temporal abstractions, spotlighting critical states in subtrajectories, and state-action abstractions, which compress the original state space and represent subtrajectories as transitions in an abstract domain. 
Building on state-action abstraction, this work aligns with ideas proposed by \cite{Dietterich1999HierarchicalRL,Ravindran2003SMDPHA,Abel2020ValuePS}, as well as the Reward Machines paradigm \citep{icarte2022reward}.

This work introduces a novel hierarchical, multi-layer framework named MANGO. Its principal aim is to learn effectively in complex, sparse-reward environments by exploiting a structured decomposition of the environment.
In addition to improving learning efficiency, a key objective is to enhance the intrinsic interpretability of the model. 
By decomposing the agent’s decision-making process across multiple levels, the user can track which subtask or higher-level goal influences the agent’s actions at each timestep. 
This level of transparency can be invaluable in industrial or safety-critical domains, where understanding an agent’s rationale and, when not aligned with the objective, correcting it is as important as its overall performance.


\section{Background}
\label{sec:background}

The concept of planning in a hierarchy of abstraction states has always been a topic of interest in the field of artificial intelligence \cite{Sacerdoti1974PlanningIA}. This approach to planning involves breaking down a complex task into a series of smaller, more manageable subtasks, which are then planned and executed independently at different levels of abstraction. 

The most commonly used family of hierarchical reinforcement learning methods for this approach is the temporal difference methods. 
These methods involve breaking down a task into temporal lags that define specific subgoals, which can then be grouped to form a macro subgoal. This allows the agent to learn to identify and achieve intermediate objectives that help to achieve the main objective, such as in \cite{sutton1999between}, \cite{Silver2012CompositionalPU}, and \cite{Konidaris2009EfficientSL}.  

In \cite{abel2022theory}, a review of the literature is presented, highlighting the main differences and characteristics of the two approaches. 
In recent years, significant research has been carried out on these topics, as demonstrated in \cite{Rafati2018LearningRI}, \cite{Konidaris2009EfficientSL}, and \cite{Konidaris2015SymbolAF}. 
However, it is worth noting that the two types of abstraction are not entirely distinct, and there are cases where they interact in state-action abstraction, as described in \cite{Dietterich1999HierarchicalRL}, \cite{Ravindran2003SMDPHA}, and \cite{Abel2020ValuePS}. 

The primary concept presented in these papers involves decomposing the Markov Decision Process (MDP) into a hierarchical structure of smaller MDPs, which are defined by specific state aggregations utilizing state abstraction. These MDPs subsequently define subtasks of the original MDP. 
In the work \cite{Abel2020ValuePS}, a mathematical formulation of state and action abstractions is provided, wherein they introduce a novel method for combining state abstractions with abstract actions, also called options \cite{sutton1999between}, by grouping them through a bisimulation metric.

We present a framework for hierarchical reinforcement learning that builds on the work of \cite{Abel2020ValuePS} and \cite{Dietterich1999HierarchicalRL}. 
Our framework uses functions that can synthesize and abstract features of the environment in a hierarchical fashion. Specifically, each layer of the hierarchy refers to a specific feature-state space, and the framework is able to discover and refine options that are defined as abstract state modifications in a particular layer of the state space.
To facilitate the learning process, we first perform a prior task-independent discovery of the options of the environment, as suggested in \cite{Veeriah2021DiscoveryOO}. We then define a reinforcement learning environment where the agent learns to perform options (macro-actions on the environment) rather than atomic actions.

In summary, our framework provides a hierarchical approach to reinforcement learning that allows for the synthesis and abstraction of environment features, discovery of options, and efficient learning through the use of macro-actions.


\section{Mathematical Framework}
\label{sec:Mathematical-Framework}

\subsection{Preliminaries}
\label{subsec:2.2}

A Markov Decision Process (MDP) is a mathematical framework for modeling decision-making in stochastic environments. 
An MDP is formally defined by the tuple ${(\mathcal{S}, \mathcal{A}, \mathcal{T}, R, \gamma)}$, consisting of a state space $\mathcal{S}$, a action space $\mathcal{A}$, a transition function ${\mathcal{T}\colon \mathcal{S} \times \mathcal{A} \to \mathbb P(\mathcal{S})}$, a reward function ${R\colon \mathcal{S} \times \mathcal{A} \times \mathcal{S} \to \mathbb{R}}$ and a discount factor $\gamma \in [0, 1)$.

An MDP captures how an agent interacts with the environment over time: at each step, the agent observes a state $s \in \mathcal{S}$, takes an action $a \in \mathcal{A}$, transitions to a new state $s' \sim \mathcal{T}(s, a)$, and receives a reward $r = R(s, a, s')$. To conclude, the goal is to learn a policy $\pi(a|s)$ which aims at maximizing the expected return, i.e. discounted sum of rewards $J$:
\begin{equation}
\label{eq:expected return}
J = \mathbb{E}\left[\sum_{t=0}^T \gamma^t r_t\right].
\end{equation}

Options, introduced by \cite{sutton1999between}, provide a structured mechanism for integrating abstract behaviors in the agent's decision process.
Options are learned sub-routines that can be executed in the environment, and can be used by the agent in an analogous way to actions, essentially expanding the action space of the agent.

Traditionally, options consist of a policy guiding agents within states, an initialization condition defining in which states the option can be initiated, and a termination condition defining when the option ends. 
In our framework, we simplify the definition, omitting the initialization condition, assuming that options can be initiated in any state of the environment.
\begin{definition}[Options]
    \label{def:option}
    Options consist of two components $o \colon = \langle \pi_o, \beta_o \rangle$:
    \begin{itemize}
        \item a policy $\pi_o \colon \mathcal{S} \to \mathbb{P}(\mathcal{A})$ guiding actions within states,
        \item a condition $\beta_o \colon \mathcal{S}\times\mathcal{A}\times\mathcal{S} \to \mathbb \{False,True\}$ defining if the option should terminate after a transition. 
    \end{itemize}
\end{definition}

The \textit{option space} is defined as the set of available options:
$$
\mathcal{O} = \{ o_1, o_2, \dots, o_k \},
$$
where each \( o_i \) is an option as previously defined. 
This expands the action space of the agent from primitive actions \(\mathcal{A}\) to a combined action-option space \(\mathcal{A}' = \mathcal{A} \cup \mathcal{O}\).

Consequently, the agent's policy $\pi$ becomes hierarchical, mapping states to distributions over this expanded space:
\begin{equation}
\label{eq:traditional agent policy}
\mu \colon \mathcal{S} \to \mathbb{P}(\mathcal{A}'),
\end{equation}
thus enabling the agent to choose either a primitive action or initiate an option at each step. When an option \(o \in \mathcal{O}\) is selected, the agent follows the option’s internal policy \(\pi_o\) until the termination condition \(\beta_o\) evaluates to \textit{True}, after which the agent resumes selecting from the expanded action-option space.

\subsection{Options as Actions in Abstract Graph}
\label{subsec:2.4}

\begin{figure}
    \label{fig:layered hierarchy}
    \begin{center}
        \includegraphics[width=\columnwidth]{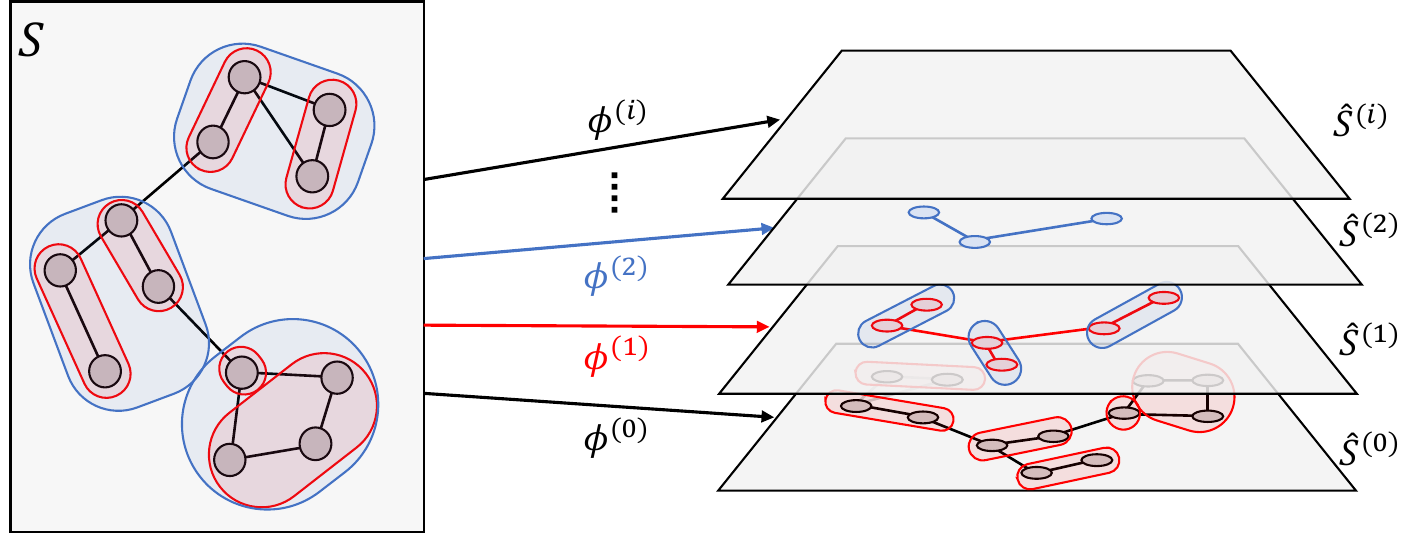}
        \caption{Progressive compression of the environment graph given a hierarchical abstraction.}
    \end{center}
\end{figure}

The state space $S$ and the transition function $\mathcal{T}$ define, respectively, the nodes and edges of a directed graph.
In order to obtain an abstract, simplified representation of this graph, we introduce the notion of Concept Function.

\begin{definition}[Concept Function]
    \label{def:concept_function}
    A concept function is a function denoted by ${\phi: \mathcal{S} \to \mathcal{S}^{\phi}}$, where $\mathcal{S}^{\phi}$ is a connected partition of the state space $S$, that is, the induced sub-graphs are connected graphs.
\end{definition}

A Concept Function should capture salient environmental attributes in a coarse, abstract representation $\mathcal{S}^{\phi}$. 
We insist that the partitions induced by concept functions must be connected. This prerequisite is vital for eliminating ambiguity in an agent's movements within the environment.  In this study, we presume that prior knowledge allows the derivation of these concepts, enabling their direct manual definition.

In an MDP, an action $a$ is represented by the transition dynamics through the transition function $\mathcal{T}_a$.
With the same objective in mind, we define an abstract action as $a^{\phi}$ to be represented by a transition function $\mathcal{T}_{a^{\phi}} \colon \mathcal{S}^{\phi} \to \mathbb{P}(\mathcal{S}^{\phi})$. The abstract action $a^{\phi}$ represents the target transition dynamic that the agent aims to achieve in the abstract state space $\mathcal{S}^{\phi}$.

Similarly to actions, options can be associated with an empirical transition function ${\mathcal{T}_{o}(s, s')}$, which describes the probability that option $o$ ends in state $s'$ given that it was initiated in state $s$. The multi-step trajectory performed by an option $o$ can be interpreted as one-step in the abstract state space $\mathcal{S}^{\phi}$.

Consequently, we establish a one-to-one correspondence between abstract actions and options, denoting each option characterized by a specific abstract action $a^{\phi}$ as $o_{a^{\phi}}$. For this purpose, we will formalize the aim of options as minimizing the dissimilarity between the target transition $\mathcal{T}_{a^{\phi}}$ and the empirical transition $\mathcal{T}_{o_{a^{\phi}}}$.


A single concept function can be associated with multiple actions, discerning the abstract action space 
$$
{\mathcal{A}^\phi=\{a^{\phi}_1, a^{\phi}_2, \dots, a^{\phi}_k \}}.
$$
This, in turn, defines the corresponding option space\footnote{To simplify the discussion, we only consider the case where the number of options is the same for all concepts, but in principle, each concept could be associated with a different number of options.} 
$$
{\mathcal{O}^\phi=\{o_{a^{\phi}_1}, o_{a^{\phi}_2}, \dots, o_{a^{\phi}_k} \}}.
$$

A Concept Function $\phi$ can also be used to naturally define a termination condition $\beta^\phi$, which triggers when the concept changes, i.e., when the agent transitions from one abstract state to another. 
This termination condition is shared across all options in $\mathcal{O}^\phi$ and can be formally defined as:
\begin{definition}[Concept Termination condition]
    \label{def:concept_termination_function}
    Given a concept function $\phi$, the termination condition $\beta^\phi$ is defined as:
    \begin{equation}
        \beta^\phi(s, a, s') = \begin{cases}
            True & \text{if } \phi(s) \neq \phi(s')\\
            False & \text{otherwise}
        \end{cases}
    \end{equation}
\end{definition}

\subsection{Structured Options using Hierarchical Abstractions}
\label{subsec:2.5}

Traditionally, options are defined directly in terms of atomic environment actions. However, allowing options to be defined in terms of other options can lead to a more structured and efficient learning process. 
In our framework, we organize options in a multi-layer hierarchy, in order to obtain a structured and well-behaved option dependency graph.

\begin{definition}[Hierarchical Abstraction]
    \label{def_AbsFunc}
    A set of concept functions ${\phi_0, \phi_1, \dots, \phi_n}$ defines a hierarchical abstraction if the respective state space partitions $\mathcal{S}^{\phi_i}$ are progressive refinements of one another. 
    That is, ${\mathcal{S}^{\phi_i} \supset \mathcal{S}^{\phi_{i+1}}}$ for all $i\in(0, 1, \dots n-1)$.
    \\
    A concept function that is part of a hierarchical abstraction will be denoted as an Abstraction Function.
\end{definition}

The refinement relation is a partial order in the set of all partitions, therefore not all combination of concept functions can be organized as hierarchical abstractions.

The identity function is the most trivial example of a concept function, where the state space is partitioned into singletons. 
This is useful for defining the base layer of the hierarchy, where the agent can perform atomic actions in the environment. 
Therefore, we define the base layer of the hierarchy $\phi_0$ to be the identity, meaning that $\mathcal{S}^{\phi_0} = \mathcal{S}$.

It is also convenient to slightly modify the shared termination condition $\beta^{\phi_0}$ to be always true, instead of triggering only when the state changes. 
This way, options at this layer of abstraction are directly connected to atomic actions in the environment.  

By requiring that partitions induced by abstraction functions are refinements of one another, we ensure that edges in higher-order graphs can be decomposed into multiple movements between lower-level states. 

This allows options at a higher abstraction level $i$ can be defined in terms of options at lower levels $j < i$. 
That is, the policy of a high-level option can invoke lower-level options as its actions, resulting in a nested, hierarchical structure. This recursive composition enables complex behaviors at higher layers to be constructed from simpler, reusable subroutines at lower layers, facilitating efficient learning and transfer across tasks.

\begin{definition}[Intra-layer Policy]
\label{policyMDPai}
An \textit{intra-layer policy} for the option ${o\in \mathcal{O}^{\phi_i}}$ is formally defined as
$$
\hat{\pi}_{o^{\phi_i}}\colon \mathcal{S} \to \mathbb{P}(\mathcal{O}^{\phi_{i-1}})
$$
and defines how options at the lower layer of abstraction can be combined to obtain the desired behaviour. 
\end{definition}

The intra-layer policy $\hat{\pi}_{o^{\phi_i}}$ implicitly defines routine associated to the option $o^{\phi_i}$, just like the agent policy in \ref{eq:traditional agent policy} is used to express complex agent behavior in terms of simpler options.
The option routine is obtained by selecting an option in $\mathcal{O}^{\phi_{i-1}}$. This lower-level option is followed until the termination condition $\beta^{\phi_{i-1}}_t$ is achieved, i.e., a state transition is executed at the lower abstraction layer $\mathcal{S}^{\phi_{i-1}}$.
After the transition is completed, once again the policy $\hat{\pi}_{o^{\phi_i}}$ provides a probability distribution that leads to the selection of the next option.




\subsection{Task actions}
\label{subsec:2.6}

\begin{figure}
    \label{fig:mango_policy}
    \begin{center}
        \includegraphics[width=0.7\columnwidth]{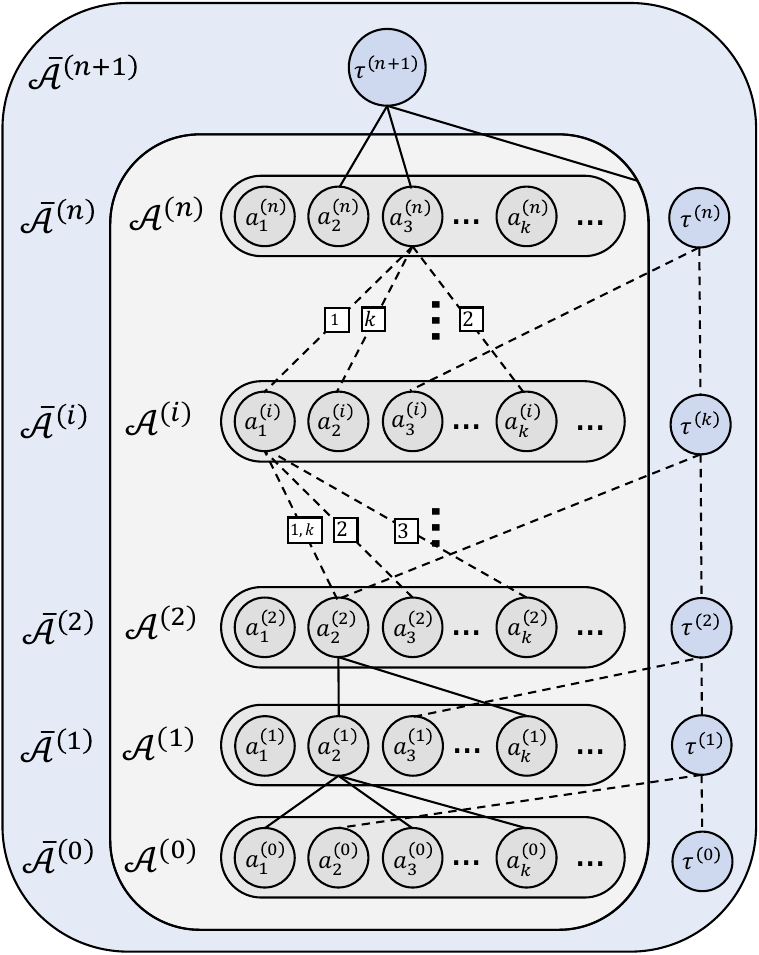}
    \end{center}
    \caption{Schematic representation of the abstract action sets and their interconnection, including the task options at each level of abstraction $\tau^{\phi_i}$.}
\end{figure}

Solving a particular task in hierarchical RL amounts to maximizing the expected return in Equation \ref{eq:expected return} accumulated following the agent policy $\mu$.

Unfortunately, given $n$ layers of abstractions $\phi_0, \phi_1, \dots, \phi_n$, each with option set of size ${| \mathcal{O}^{\phi_i} | = k}$, yields a combined options set of size $(n+1) \cdot k$. 
Thus, the size of the option set obtained using the proposed framework grows relatively rapidly as more and more abstraction layers are considered, making the direct optimization of the policy $\mu$ increasingly difficult. 
However, the agent policy $\mu$ can be restructured in a way that takes advantage of the hierarchical organization of the option space: instead of a single policy over a large option space, a hierarchy of simpler policies is used.

Similarly to the hierarchical organization of options, the agent's policy can be arranged within the hierarchy as an additional abstract action integrated to each layer with an additional instantiation of an upper layer $(n+1)$ that will move the agent using the highest level options.

\begin{definition}[Task Actions and Policies]\label{def:task_action_policy}
A \emph{task action} $\tau^{\phi_i}$ is an abstract action defined at any layer $i \in \{0,\dots,n,n+1\}$. The corresponding expanded abstract action space is given by $\bar{\mathcal{A}}^{\phi_i} := \mathcal{A}^{\phi_i} \cup \{\tau^{\phi_i}\}$.

Each task action $\tau^{\phi_i}$ induces a corresponding \emph{task policy}, denoted by $\pi_{\tau^{\phi_i}}: \mathcal{S} \to \mathbb{P}(\bar{\mathcal{A}}^{\phi_{i-1}})$, which operates over the expanded action space of the layer below. These policies, analogous to those associated with other abstract actions, capture the hierarchical structure of decision-making. Due to the temporal abstraction introduced by this recursive structure, task policies are defined as \emph{semi intra-layer policies} $\hat{\pi}_{\tau^{\phi_i}}$.
\end{definition}

While task actions mirror abstract actions structurally, they differ in termination: abstract actions end on an abstract state change, whereas task options terminate without inducing any state change at layer $\phi_i$ Execution cascades upward, allowing higher-layer intra-layer policies to resume execution, and terminates upward, layer by layer, until reaching the top-level option.

The resultant subtrajectories are then linked either to previously trained options characterized by abstract actions $a^{\phi_{i-1}}$ or to task options $o_{\tau^{\phi_i}}$ if identified by $\tau^{\phi_{i-1}}$. 
The termination function.
\begin{definition}[Task option termination function]\label{def:task_term_funct}
Given the task option $\tau^{\phi_i}$, the termination function $\beta^{\tau^{\phi_i}}$ is defined as:
$$
\beta_{\tau^{\phi_i}}(s_{t},\bar{a}^{\phi_{i-1}}_{t}, s_t) = 
\begin{cases}
\text{True} \text{ if  } \bar{a}^{\phi_{i-1}}_{t} = \tau^{\phi_{i-1}}\\
0 \text{ otherwise }\\
\end{cases}
$$
except for $o_{\tau^{(0)}}$, which is the null option (empty routine)
\end{definition}

\subsection{Learning Process}
\label{subsec:2.7}

We defined a generic option $o_{\bar{a}^{\phi_i}}$ associated with the action $\bar{a}^{\phi_i} \in \bar{\mathcal{A}}^{\phi_i}$ as a routine in the base layer, where $\bar{a}^{\phi_i}$ is an abstract action or a task action. 
This option routine comprises a chain of semi-intra-layer policies, which navigate through the states of the $(j)$-th level of abstraction with $j<i$ until the termination conditions are encountered.

An MDP in the layer of abstraction $\phi_i$ consists of the set of states $\mathcal{S}$ introduced in Section \ref{sec:background}, the abstract actions $\mathcal{A}^{\phi_i}$ introduced in Section \ref{subsec:2.2}, the group of transition function $\mathcal{T}_{a^{\phi_i}}$ with $a^{\phi_i} \in \mathcal{A}^{\phi_i}$ introduced in Section \ref{subsec:2.2} and a specific discount factor $\gamma\phi_i$. Given that the MDP provide the structure to train the specific policy that replicates the trajectory within layer $\phi_i$ when executing an action $a^{\phi_{i+1}}$ from the layer above,we charactewrize the reward associated to this specific action defining it by $R_{a^{\phi_{i+1}}}$.
Since there are multiple actions from the layer above $a^{\phi_{i+1}} \in \mathcal{A}^{\phi_{i+1}}$, corresponding to different options $o_{a^{\phi_{i+1}}}$ with their intra-layer policies in layer $\phi_i$, we create distinct MDPs within layer $\phi_i$ with identical structural elements $(\mathcal{S},\mathcal{A}^{\phi_i},\{\mathcal{T}_{a^{\phi_i}}|a^{\phi_i} \in\mathcal{A}^{\phi_i}\}, \gamma\phi_i )$ but varying reward functions aligned with the abstract actions from the layer above.


\begin{definition}[SMDP of the action $a^{\phi_i}$]
\label{smdp}
Given the two Layer of a Hierarchical Abstraction $\phi_i$ and $\phi_{i+1}$, given an action $\bar{a}^{\phi_{i+1}} \in \bar{\mathcal{A}}^{\phi_{i+1}}$ we define the Markov Decision Process of that action as the tuple of: structural objects, the discount factor and the reward functions that shape the environment of the layer $\phi_i$ according to given action $\bar{a}^{\phi_{i+1}}$,  $(\mathcal{S}, \bar{\mathcal{A}}^{\phi_i}, \{\mathcal{T}_{\bar{a}^{\phi_i}}|\bar{a}^{\phi_i} \in\bar{\mathcal{A}}^{\phi_i}\}, \gamma\phi_i, R_{\bar{a}^{\phi_{i+1}}})$. 
${R_{\bar{a}^{\phi_{i+1}}}\colon S \times \bar{\mathcal{A}}^{\phi_i} \times S \to \mathbb{R}}$  is the reward function such that $R_{\bar{a}^{\phi_{i+1}}}(s_t, a^{\phi_i}_{t},s_{t+1})$ is given by:
\begin{equation}
 \sum_{(s_{k}, a_k,s_{k+1}) \sim o_{\bar{a}^{\phi_i}_t}} \gamma_{\phi_i}^{k} r_{\bar{a}^{\phi_{i+1}}}(s_{k}, a_k, s_{k+1})
\end{equation}
and $r_{\bar{a}^{\phi_{i+1}}}(s_{k}, a_k, s_{k+1})$ is defined as:
\begin{multline}
\begin{cases} R(s_{k}, a_k, s_{k+1})  \\
\;\;\;\;\;\;\;\;\;\;\;\;\;\;\;\;\;\;\;\;\text{ if  }  
  \bar{a}^{\phi_{i+1}}=\tau^{\phi_{i+1}}\\ 
\beta^{\phi_{i+1}}_k(1 - 2 d_{\mathcal{T}_{a^{\phi_{i+1}}}}(\delta{(\phi^{\phi_i}(s_k),\phi^{\phi_i}(s_{k-1}))})) \\
\;\;\;\;\;\;\;\;\;\;\;\;\;\;\;\;\;\;\;\;\text{ if  } \bar{a}^{\phi_{i+1}}=a^{\phi_{i+1}}
\end{cases}
\end{multline}
\end{definition}

The function \(r_{\bar{a}^{\phi_{i+1}}}(s_{k}, a_k, s_{k+1})\) takes into consideration the current state \(s_k\), the action \(a_k\) taken, and the subsequent state \(s_{k+1}\), all contingent on the macro action \(\bar{a}^{\phi_{i+1}}\) at hierarchy level \(i+1\).
When \(\bar{a}^{\phi_{i+1}}\) corresponds to a task action \(\tau^{\phi_{i+1}}\), the reward is derived from the standard reward function \(R(s_{k}, a_k, s_{k+1})\).
However, if \(\bar{a}^{\phi_{i+1}}\) represents an abstract action \(a^{\phi_{i+1}}\), the reward is intrinsic in the structure and influenced by the abstract transition generated by the trajectory. It is nonzero only if the trajectory terminates, which is indicated by \(\beta^{\phi_{i+1}}_{k+1}=1\). Additionally, the reward is adjusted based on the alignment of the occurred abstract transition with the intended transition defined by the abstract action dynamics, quantified by \((1-2 d_{\mathcal{T}_{a^{\phi_{i+1}}}}(\delta{(\phi^{\phi_i}(s_k),\phi^{\phi_i}(s_k-1))}))\).

It is then possible to explicitly write the Q-function related to the training of the policy of the an option at a given layer $\phi_i$ as \( Q_{\bar{a}^{\phi_i}}(s_t, o_{\bar{a}^{\phi_{i-1}}_t}) \) where  $\bar{a}^{\phi_{i-1}}_t \in \bar{\mathcal{A}}^{\phi_{i-1}}$:

\begin{multline}
Q_{\bar{a}^{\phi_i}}(s_t, \bar{a}^{\phi_{i-1}}_t) = \mathbb{E}_{s_{t+1} \sim \pi_{o_{\bar{a}^{\phi_{i-1}}_t}}}[ r_{\bar{a}^{\phi_i}}(s_t, 
\bar{a}^{\phi_{i-1}}_t,s_{t+1}) + \\
\gamma^{\phi_{i-1}} \cdot \max_{\bar{a}^{\phi_{i-1}}_{t+1} \in \bar{\mathcal{A}}^{\phi_i}} Q(s_{t+1}, \bar{a}^{\phi_{i-1}}_{t+1})]
\end{multline}

Here, \( Q_{\bar{a}^{\phi_i}}(s_t, \bar{a}^{\phi_{i-1}}_t) \) represents the expected cumulative reward gained by the policy $\bar{a}^{\phi_i}$ starting from state \( s_t \) and executing the option characterized by the action \(\bar{a}^{\phi_{i-1}}_t \). The expectation is taken over the state distribution \( s_{t+1} \) at the end of the policy characterized by the action $\bar{a}^{\phi_{i-1}}_t$. The expression \( r_{\bar{a}^{\phi_i}}(s_t, o_{\bar{a}^{\phi_{i-1}}_t},s_{t+1}) \) denotes the reward shaped by $\bar{a}^{\phi_i}$ obtained by executing option \(o_{\bar{a}^{\phi_{i-1}}_t} \) at state \( s_t \), and \( \gamma^{\phi_i} \cdot \max_{\bar{a}^{\phi_{i-1}}_{t+1} \in \bar{\mathcal{A}}^{\phi_i}} Q(s_{t+1}, \bar{a}^{\phi_{i-1}}_{t+1}) \) is the discounted maximum expected future reward from the next state \( s_{t+1} \), considering the abstract actions available for the subsequent step $\bar{\mathcal{A}}^{\phi_i}$.

After training various task actions at each layer, the top-level task option $\tau^{\phi_{i+1}}$ can then act as the agent's policy, represented as $\mu$. This policy oversees the entire process, utilizing the options to orchestrate the agent's movements and terminates with the conclusion of the environment's entire process.


\section{Experimental Setup and Results}
\label{sec:experiments_results}

\subsection{Environment Description}
\label{sec:Environment}

We adopt a grid-based environment loosely inspired by Frozen Lake, where each cell can be frozen (safe to walk on) or a hole (causing an immediate episode termination). The agent starts in one cell, and the goal (or “gift”) is placed in another. The agent earns a reward upon reaching this goal, and each episode terminates either when the agent succeeds (reaching the goal) or fails (falling into a hole or exceeding a step limit).
Once generated, the map remains fixed, but the agent and the goal object (“gift”) are placed randomly at each reset.

To apply MANGO, we define a hierarchy of abstraction functions $\phi_i$ that compress contiguous regions of the grid into “supercells.” For instance, at level $i$, each $(2^i \times 2^i)$-block of the original map collapses into a single abstract cell, preserving only the essential one-hot information for the agent’s position and goal. By ensuring connectedness and consistent transitions, these abstractions meet the criteria in Section \ref{subsec:2.4}.

At each layer $i$, we define the movement actions in the compressed (abstract) grid \texttt{\{UP,LEFT,DOWN,RIGHT\}}. An abstract action thus corresponds to relocating the agent from one abstract cell to an adjacent one at layer $i$, with collisions or holes folded into that layer’s transition dynamics.

\begin{figure}
    \begin{center}
        \includegraphics[width=\columnwidth]{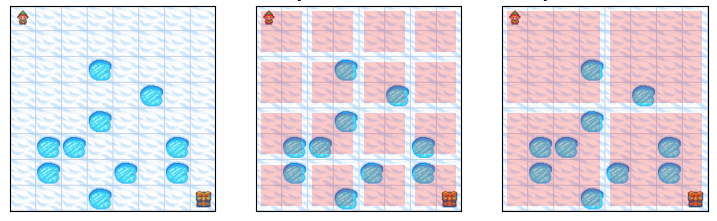}
        \caption{Abstraction functions applied to the Frozen Lake environment}
        \label{fig:abstractions}
    \end{center}
\end{figure}

Figure \ref{fig:abstractions} illustrates how blocks of the map are pooled at each layer, visually conveying which states are aggregated and how transitions map between adjacent abstract states.

\subsection{Training Procedure}\label{sec:training_procedure}

A set of parallel workers then collects environment rollouts by interacting with the map. These transitions, including states, rewards, and terminations, serve as input for off-policy RL updates.

The first (lowest) layer of MANGO trains for a prescribed number of steps, refining a policy that learns basic navigation and the ability to avoid holes. When this layer’s budget is exhausted or its performance is deemed satisfactory, its parameters are frozen, meaning that future layers can treat it as part of the environment’s transition model of the upper layer. The next layer operates on a more abstract representation of the map, delegating detailed actions to the frozen sub-policy. This process continues until all hierarchical layers have undergone training, each refining a policy at its own level of abstraction.

Multiple objective functions are tested to balance early improvements against ultimate convergence. These include maximizing the area under the reward-step curve, combining final accuracy with training length, and rewarding both fast and stable convergence. In practice, the method proves highly sensitive: if the first layer is not adequately trained, its errors propagate to subsequent layers, preventing the outer policy from achieving full accuracy. Conversely, overly aggressive learning rates can produce rapid initial gains but hamper fine-tuning. 

\subsection{Results}
\label{sec:results}

\begin{figure}
    \begin{center}
        \includegraphics[width=0.8\columnwidth]{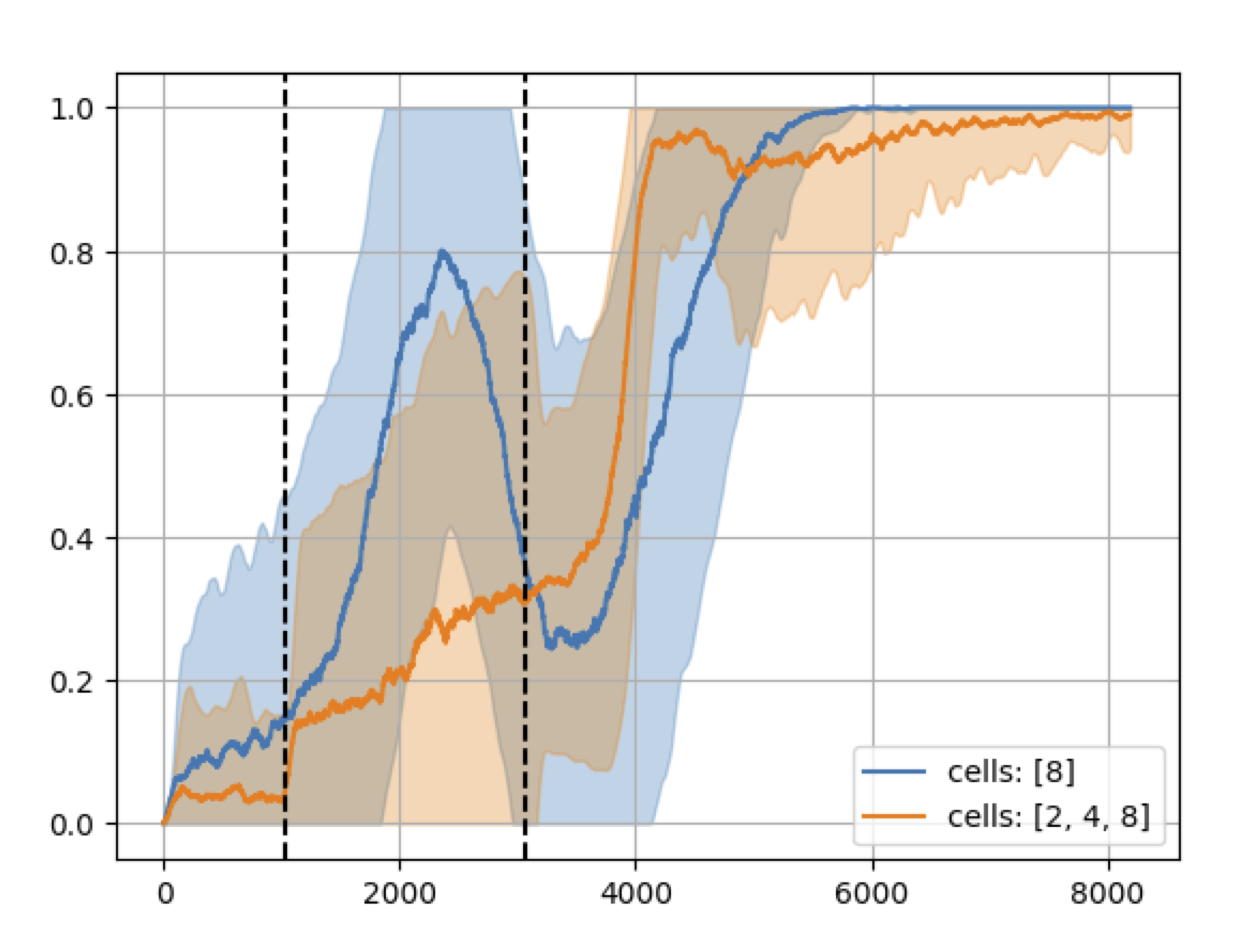}
        \caption{Convergence of MANGO and Q-learning with hyperparameter optimization, the vertical dashed lines represent when MANGO starts the training of the upper layer. We can observe that mango reaches the first high percentage but saturates later than Q-learning.}
        \label{fig:mangotraining}
    \end{center}
\end{figure}

\begin{figure}
\centering
    \begin{center}
        \includegraphics[width=0.6\columnwidth]{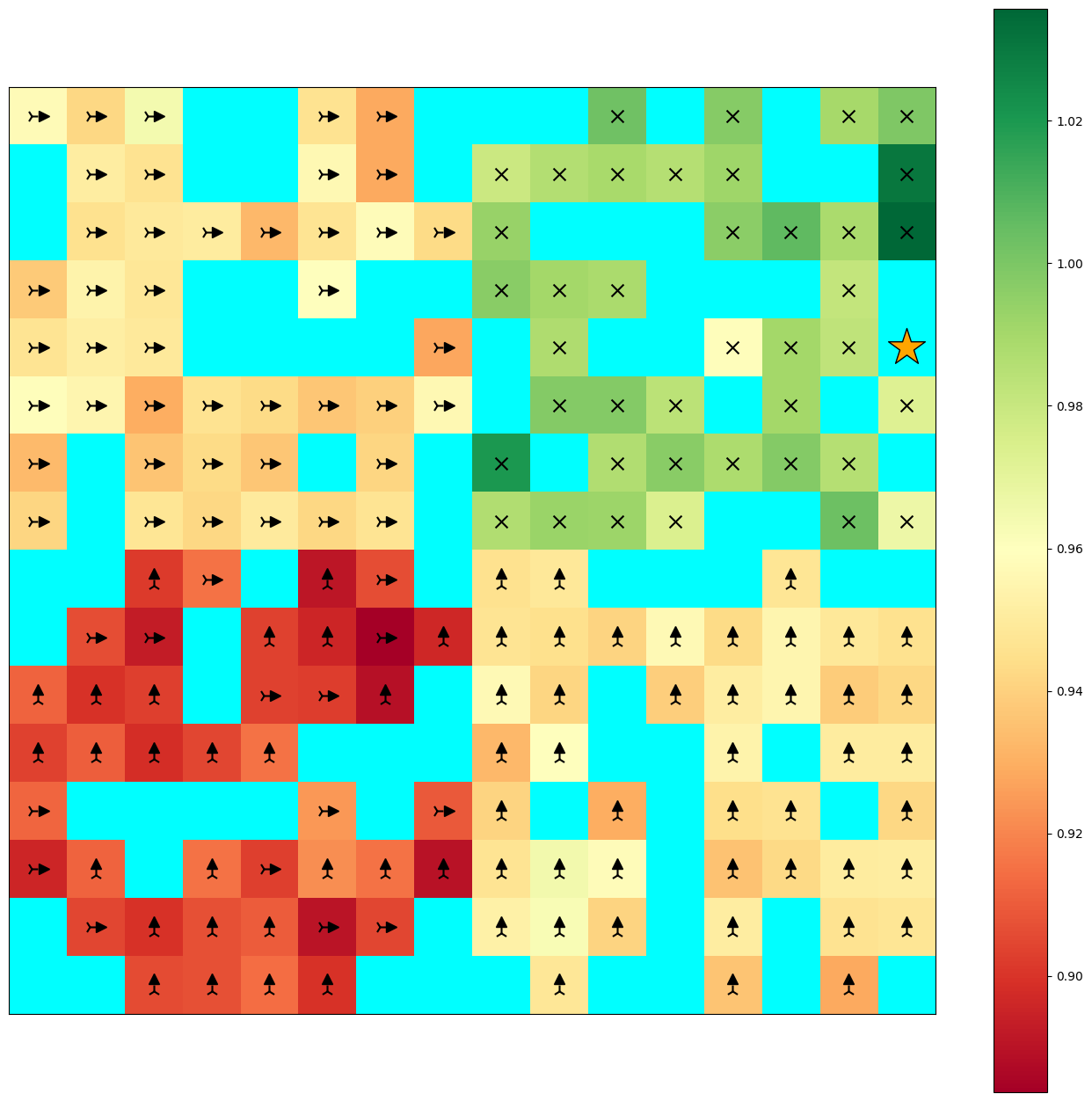}
    \end{center}
    \caption{Q-values of the overarching task action at level 4 of abstraction.}
    \label{fig:task_layer}
\end{figure}


The hierarchical training strategy described in Section~\ref{sec:training_procedure} can successfully guide the agent on $16 \times 16$ randomly generated Frozen Lake maps, provided the hyperparameter configuration is carefully tuned. In many trials, the agent achieves near-perfect completion rates once each layer’s policy has been given enough training steps to master its designated abstraction as shown in Figure \ref{fig:mangotraining}. However, even slight misconfigurations in learning rates, soft-update schedules, or layer-wise training budgets often lead to partial convergence or complete failure. 

One consistent observation is that if the base layer does not thoroughly learn to navigate local obstacles and avoid holes, higher layers inherit an unreliable subroutine for their coarse abstractions, causing the entire policy stack to stall below optimal performance. In contrast, when the first layer converges robustly, subsequent layers can assemble its transitions into longer, more abstract paths that accelerate exploration and convergence. Yet this hierarchical advantage vanishes under overly aggressive learning rates, which may yield rapid initial gains at lower layers but prevent fine-tuning; Optuna’s pruning can also prematurely end otherwise promising runs that start slowly.


\section{Conclusion}

This work introduced MANGO, a hierarchical reinforcement learning framework that decomposes complex environments into layers of abstract state transitions and nested options. Through this structure, MANGO improves sample efficiency and interpretability by enabling agents to reason and act at multiple levels of abstraction. Experiments in sparse-reward grid environments demonstrate the framework's effectiveness when properly tuned, although performance remains sensitive to the quality of lower-layer policies and training schedules.

Beyond improved learning efficiency, MANGO enhances interpretability by making the agent’s decision-making process transparent across layers—an asset in safety-critical or industrial settings. Future work will explore automated discovery of abstractions and abstract actions, adaptation to continuous or fuzzy environments, and more robust multi-layer training strategies. These directions aim to strengthen MANGO’s scalability, stability, and autonomy in increasingly complex tasks.


\bibliography{ifacconf}
\appendix

\end{document}